\documentclass{article}

\usepackage{arxiv}

\usepackage[utf8]{inputenc} 
\usepackage[T1]{fontenc}    
\usepackage{hyperref}       
\usepackage{url}            
\usepackage{booktabs}       
\usepackage{amsfonts}       
\usepackage{nicefrac}       
\usepackage{microtype}      
\usepackage{blindtext}
\usepackage{mdframed}
\usepackage{listings}
\usepackage{times}
\usepackage{graphicx}
\usepackage{latexsym}
\usepackage{cite}
\usepackage{subfig}
\usepackage{caption}
\usepackage{multicol}
\usepackage[all]{nowidow}
\usepackage{tabularx}

\usepackage{soul}
\usepackage{xcolor}

\definecolor{cb-black}{RGB}{0,0,0}
\definecolor{cb-orange}{RGB}{230,159,0}
\definecolor{cb-skyblue}{RGB}{86,180,233}
\definecolor{cb-bluishgreen}{RGB}{0,158,155}
\definecolor{cb-yellow}{RGB}{240,228,66}
\definecolor{cb-blue}{RGB}{0,114,178}
\definecolor{cb-vermillion}{RGB}{213,94,0}
\definecolor{cb-reddishpurple}{RGB}{204,121,167}

\newlength\myboxwidth
\newcommand{\etal}{\emph{et al.}}
\newcommand{\ie}{\emph{i.e.}}
\newcommand{\eg}{\emph{e.g.}}

\newcommand{\investigation}[2][?]{\smallskip\noindent\emph{\textbf{#1 - }#2}}
\newcommand{\hyql}[1]{\noindent\texttt{\footnotesize{#1}}}

\newcommand{\hyqlinline}[1]{\texttt{\small{#1}}}

\newcommand\tab[1][0.5cm]{\hspace*{#1}}

\title{Managing Machine Learning Workflow Components}

\author{
  Marcio Moreno, V\'{i}tor Louren\c{c}o, Sandro Rama Fiorini,\\\textbf{Polyana Costa, Rafael Brand\~{a}o, Daniel Civitarese, and Renato Cerqueira}\\
  IBM Research, Brazil \\
  \texttt{mmoreno@br.ibm.com, vitor.nascimento@ibm.com, sandro.fiorini@ibm.com,}\\ \texttt{polyana.bezerra@ibm.com, rmello@br.ibm.com, sallesd@br.ibm.com, rcerq@br.ibm.com} \\
}

\begin{document}
\maketitle

\begin{abstract}
Machine Learning Workflows~(MLWfs) have become essential and a disruptive approach in problem-solving over several industries. However, the development process of MLWfs may be complex, time-consuming, and error-prone. To handle this problem, we introduce \emph{machine learning workflow management}~(MLWfM) as a technique to aid the development and reuse of MLWfs and their components through three aspects: representation, execution, and creation. We introduce our approach to structure MLWfs' components and metadata in order to aid component retrieval and reuse of new MLWfs. We also consider the execution of these components within a tool. A hybrid knowledge representation, called Hyperknowledge, frames our methodology, supporting the three MLWfM's aspects. To validate our approach, we show a practical use case in the Oil \& Gas industry. In addition, to evaluate the feasibility of the proposed technique, we create a dataset of MLWfs executions and discuss the MLWfM's performance in loading and querying this dataset.
\end{abstract}


\section{Introduction}	

The recent advances in machine learning~(ML), especially in neural networks~\cite{lecun2015}, leverage capabilities of problem-solving in a broader sense,~\ie{}, being applied on cross-industries domain, varying for instance from Smart Cities and Public Security~\cite{lourenco2018,zortea2018} to Agriculture~\cite{nery2018facing,moreno2018smart} and Oil \& Gas~(O\&G)~\cite{civitarese2019}. In general, to address tasks in these domains, it is necessary to create complex ML Workflows~(MLWf). In this paper, we consider that MLWfs are composed of components (\eg{}, datasets, models, associated tasks) and the relationship between the the components. The development process of the MLWf usually produces a considerable number of components and different instances that are, as expected, task-specific. Some examples are detecting objects in an image, predicting possible links within a graph, and others. This development process may be complicated, hard to achieve, time-consuming, and error-prone~\cite{morenopatent2019}. Furthermore, the unstructured growth of MLWf limits the reuse of components, since there is no well-defined common vocabulary to structure them. To overcome these issues, an essential aspect of MLWf development, yet commonly overlooked, is the management of the machine learning workflows themselves and their components in what we call \emph{machine learning workflow management}.

In this paper, we define the Machine Learning Workflow Management~(MLWfM) as a technique where the ontology-based description of MLWfs provides the means for executing existent MLWf and the automatic creation of new ones. The ontology-oriented structuring provides a common vocabulary that helps achieving interoperability between the MLWf's components as well as supports searching for components by their characteristics. The creation of a new MLWf relies on the ML concepts described in the ontology and the available characteristics to reuse existing components to produce new MLWf in different contexts.

We argue that MLWfM can contribute to the development and reuse of MLWf through the aspects of structuring, execution, and creation. Despite the novelty of the term \emph{machine learning workflow management}, some of these issues have been individually trackled by other works in literature. However, the majority focus on execution or limited creation of MLWf~\cite{jannach2016,zaharia2018}, while some propose a common vocabulary to structure the MLWf~\cite{esteves2015,publio2018}. None addresses the three aspects within what we consider necessary to achieve MLWfM. In this paper, we present the machine learning workflow management as a technique to address the three issues in MLWf development. We propose the use of hybrid knowledge representation to structure broad ML tasks by putting together the best set of elements (specific tasks) to automatize this process. The main contributions of this paper are (\emph{i})~a new knowledge-oriented representation for MLWf; and (\emph{ii})~a framework to support the MLWfM and its execution and creation features.

To illustrate our work, we explore an industrial use case in O\&G Exploration that relies on multiple machine learning and data processing workflows. This use case presents an ontology-based representation of a MLWf and how this representation and the components' semantic metadata provide support to reuse. Additionally, we show the benefits of MLWfM in the development process of MLWf through the execution of existing components on the use case and the creation of new workflows by reusing existing components. We validate our approach showing scenarios in which we address the requirements by using the stated MLWfM. Finally, we introduce a new dataset proposed to experimentally evaluate the loading and querying response time of the MLWf.

The remainder of this paper is organized as follows: Section 2 presents the main related work; Sections 3 and 4 show the knowledge-oriented representation of a MLWf and the MLWfM, respectively; in Section 5, we provide the validation of our approach and further discussions; we introduce in Section 6 the proposed dataset and experimental evaluation over it. Finally, in Section 7, we present the final remarks and future directions.

\section{Related Work}
We discuss the related work regarding two aspects: \emph{(i)} representation of MLWfs and \emph{(ii)} the usage and reuse of MLWf components.

Addressing the gap between execution provenance of a machine learning workflow and its representation for the reproducibility of executions, Esteves~\etal{}~\cite{esteves2015} and Publio~\etal{}~\cite{publio2018} propose the use of a vocabulary and an ontology schema, respectively. The MEX vocabulary~\cite{esteves2015} provides a standard schema based on a machine-readable terminology that aids the reproducibility of execution in various frameworks and workflow systems. However, it lacks details of the machine learning process itself, focusing mainly on the general description machine learning workflow. To overcome this issue, the W3C ML-Schema~\cite{publio2018} extends the MEX vocabulary improving the representation of the machine learning process. Both approaches structure the workflows within a common vocabulary, which is a fundamental aspect of the MLWfM. However, they focus on the reproducibility of executions, partially supporting the reuse of the MLWf's components.

MLflow\footnote{https://mlflow.org/}~\cite{zaharia2018} and IBM Watson Studio\footnote{https://www.ibm.com/cloud/machine-learning} are available solutions for designing and deployment of machine learning workflows. The advantages of these systems rely on the support for the creation and modification of machine learning workflows in a stable environment. The main drawback of these approaches is the lack of a knowledge representation, which could take advantage of a common vocabulary and support MLWf components' metadata structuring.

Jannach~\etal{}~\cite{jannach2016} proposes a recommendation system plugin to RapidMiner\footnote{ttp://www.rapidminer.com}, which supports the development of machine learning workflows trough adaptive recommendations based on a predictive model trained over existing machine learning workflows. Wang~\etal{}~\cite{wang2018} propose a unified system architecture called Rifiki, that allows users of machine learning models to train and to predict through built-in services without handling specifics of building models, tuning hyper-parameters, optimization, and others. These approaches provide frameworks to ease the development of machine learning models, but they do not provide adequate representation, limiting the reuse capabilities of components.

Carvalho~\etal{}~\cite{carvalho2018} propose a semantic software catalog to aid scientists in managing their computational experiments workflow exploration and evolution. The named \textsc{OntoSoft-VFF} is built upon a new ontology, which describes software functionalities and evolution through their semantic metadata, affording to query over the represented components and its metadata. They have also illustrated their method over a machine learning workflow. Compared with other works mentioned above, Carvalho~\etal{}'s approach is more aligned with the proposed MLWfM. Nonetheless, it is focused on the representation of general software, lacking details of the MLWf. Also, it does not support the creation and execution of the workflows.

\section{Representation of Machine Learning Workflows}

In this section, we describe the knowledge model elements that support our MLWfM method. The core component is the ML Schema proposed by W3C as a core vocabulary for the machine learning domain. 

\subsection{Hyperknowledge}\label{sec:hk}

Hyperknowledge is a knowledge representation model that supports the representation of high-level semantic concepts, multimodal information, unstructured data, and the relationships between them in a unified way~\cite{moreno2016ncm,moreno2017extending,moreno2019}. Its conceptual model has the expressiveness to relate multimedia content (\emph{e.g.}~image, audio, text, video), with abstract concepts (\emph{e.g.}~label, classes) within the same framework. Besides, Hyperknowledge supports the representation of formal descriptions present in ontologies, linked data, machine learning models, executable content, and source code. By providing a flexible descriptive framework, Hyperknowledge helps to fill the semantic gap between hypermedia content and symbolic knowledge content, allowing reasoning over cross-modal types of information.

The Hyperknowledge model is composed of nodes, links, connectors, contexts, and supporting constructs. Nodes represent resources and can be decorated with properties and anchors, properties represent node characteristics with literal values, and anchors denote fragments of the resource denoted by the node. For example, an anchor on a node representing an image might denote a region on that image. All nodes have a \emph{lambda} anchor denoting the whole resource. Links can associate two or \emph{more} nodes, while the connectors specify the semantics of a link. This feature differentiates it from traditional graph-based representations, allowing the representation of $n$-ary relations without reification. Links connect to nodes exclusively through anchors. Links can also display properties like nodes.

Furthermore, Hyperknowledge graphs are organized into contexts. All nodes and links are in a context; if a context is not specified, then the default context is assumed. Contexts are composite nodes, so they can be linked themselves to other nodes and also have parent contexts, effectively allowing the representation of context hierarchies.

\subsection{ML Schema}\label{sec:mlschema}

\begin{figure*}[h!]
    \centering
    \begin{mdframed}
    \centering
    \includegraphics[width=0.9\columnwidth]{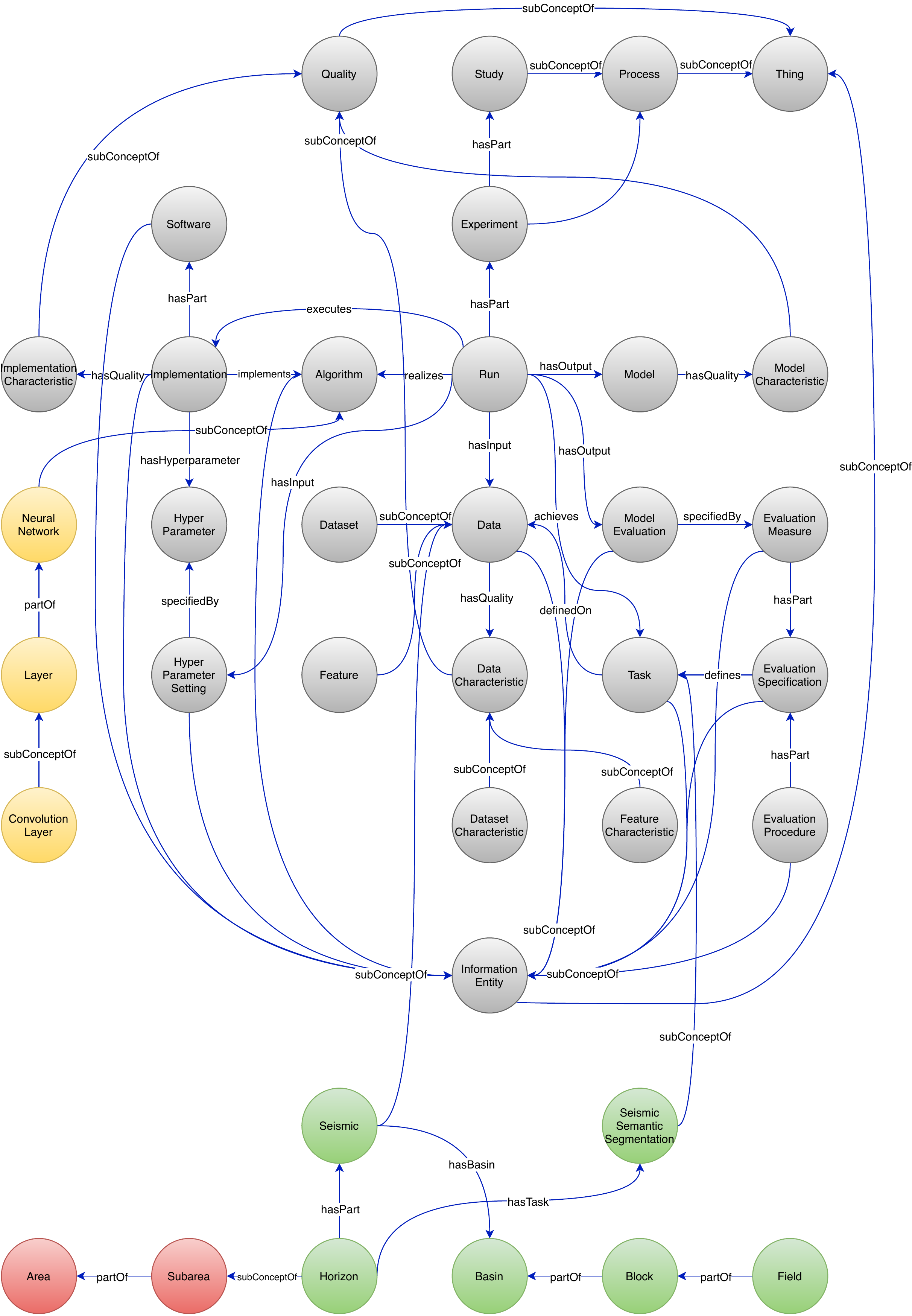}
    \end{mdframed}
    \caption{Hyperknowledge model of ML Schema vocabulary~\cite{publio2018} (in gray) and some additional concepts. Concepts in yellow, green, and red denote, respectively, specific ML concepts we specified to be used in the examples below; the domain concepts for the use case in Section~\ref{sec:use-case}; and the domain concepts necessary for the dataset proposed in Section~\ref{sec:dataset}.}
    \label{fig:full-ontology}
\end{figure*}

ML Schema defines constructs that allow one to describe ML algorithms, tasks, implementations, and executions (Figure~\ref{fig:full-ontology}). It can be used as a basis for the specification of ontologies, databases, and APIs for machine learning. We translate the OWL\footnote{https://www.w3.org/OWL/} implementation of ML Schema to Hyperknowledge to allow instantiation and query of ML models. Concepts and relations in the ML Schema OWL model are translated to Hyperknowledge nodes and links, classes are converted to nodes, and object properties are binary links between concepts. Datatype properties are properties on concept nodes, and all the nodes and their links added to a single context. We specify the ML Schema ontology in its context, while extensions and instantiations were added to separated contexts. This contextual structuring allows for a more organized knowledge model, with separation of concerns for each context.

\section{Machine Learning Workflow Management\label{sec:MLWfM}}

This section discusses the concepts regarding \emph{machine learning workflow management} and how Hyperknowledge supports these concepts, namely: \emph{(i)} MLWfs' components retrieval, \emph{(ii)} MLWf creation, and \emph{(iii)} MLWf execution.

One of the advantages of the MLWfM is the knowledge framing described in Section~\ref{sec:mlschema}. Hyperknowledge structures the MLWf within a knowledge base, which enables them to be searched and have their components retrieved for further composition. In such a way, one could search for components individually or entire workflows regarding their functionalities and metadata,\ie{}, any information referring data from components or workflows (\eg{}, execution log, parameter value). Furthermore, since the MLWfM relies on Hyperknowledge representation, it is possible to develop queries using the Hyperknowledge Query Language~(HyQL).

As an example, one possible query could be to search for all models developed to achieve classification. The structure of this query with HyQL would be:

\begin{mdframed}
\hyql{
SELECT Model WHERE\\
\tab Run achieves Classification AND Run hasOutput Model}
\end{mdframed}

However, such a query could result in a broad set of models. As discussed in Section~\ref{sec:hk}, entities can have anchors that represent part of its content. For instance, in our example, the desired models could have a convolution layer. Besides, to enhance the search providing a filtered result, the models' metadata could also be part of the query. Like so, we propose a new enhanced query:
\newpage
\begin{mdframed}
\hyql{
SELECT Model WHERE\\
\tab Run achieves Classification AND Run realizes Algorithm AND\\
\tab Algorithm\#ConvolutionLayer AND Run hasOutput Model AND\\
\tab Model.accuracy > 0.9}
\end{mdframed}

In both cases, the retrieved entities are components from already represented workflows. This approach yields the possibility of creating new MLWf from the retrieval of existing components, another advantage of using MLWfM.

\begin{figure}[h]
    \centering
    \subfloat[][]{\includegraphics[width=0.49\columnwidth]{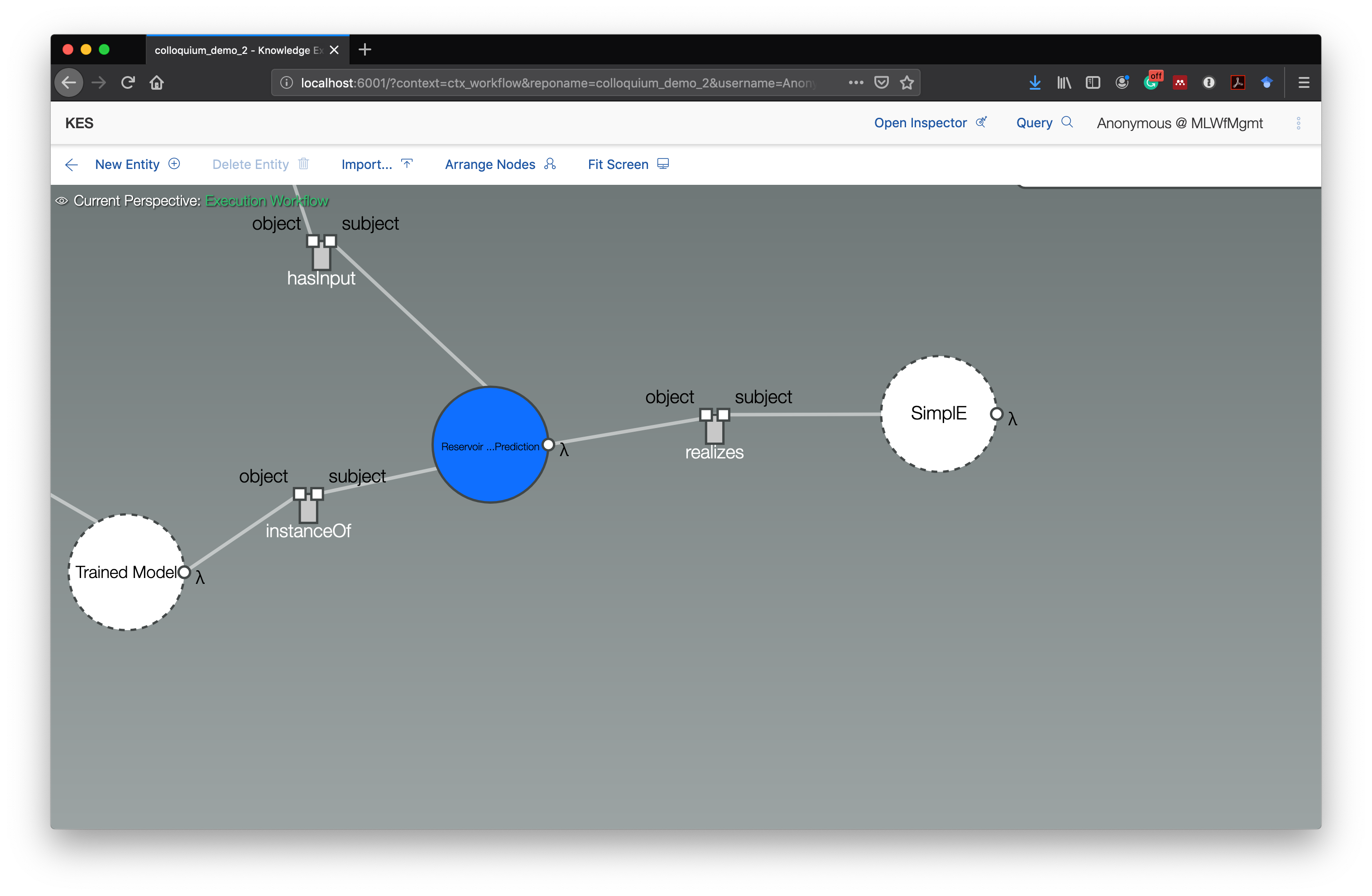}}%
    \subfloat[][]{\includegraphics[width=0.49\columnwidth]{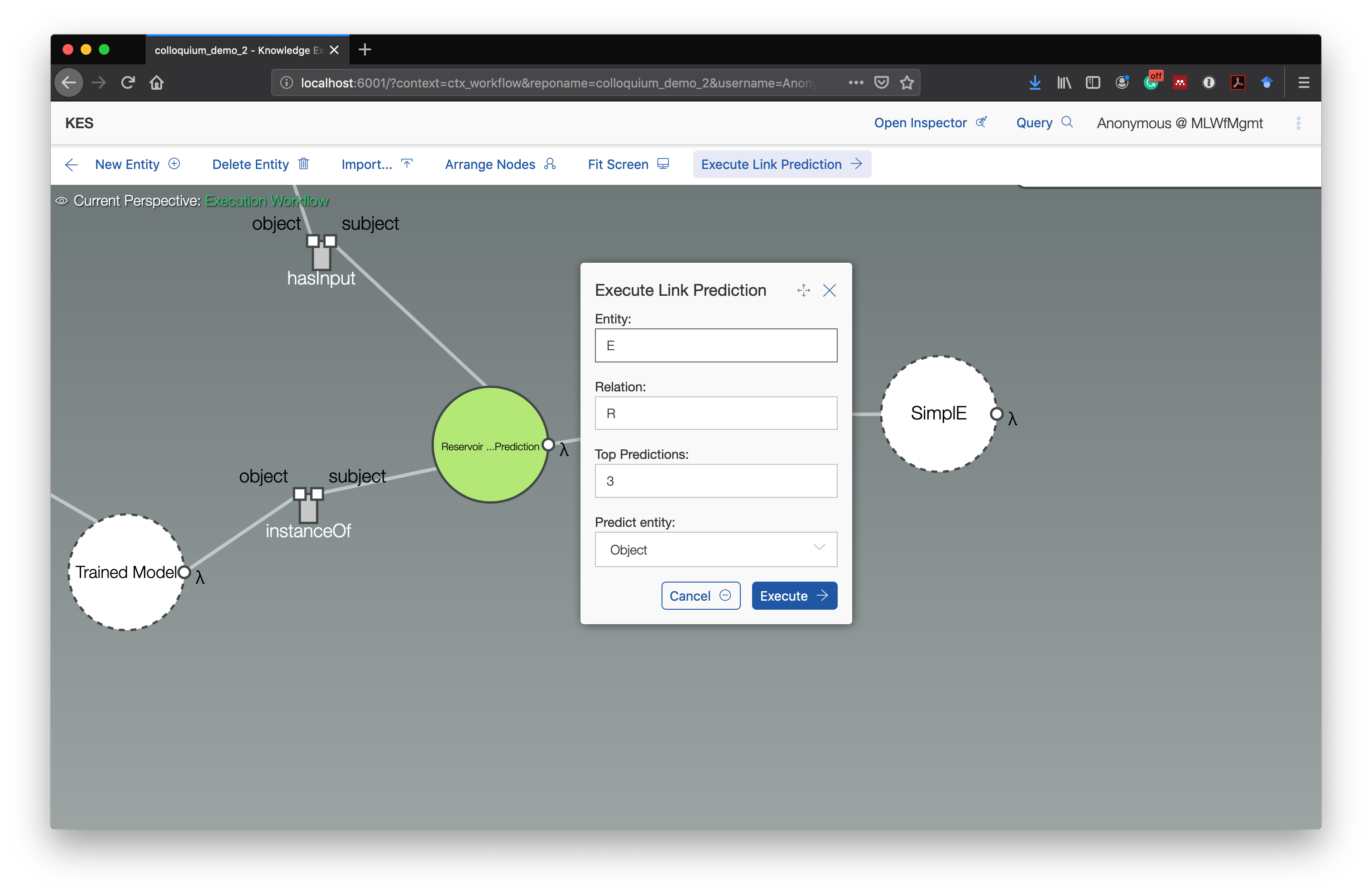}}\\%
    \subfloat[][]{\includegraphics[width=0.49\columnwidth]{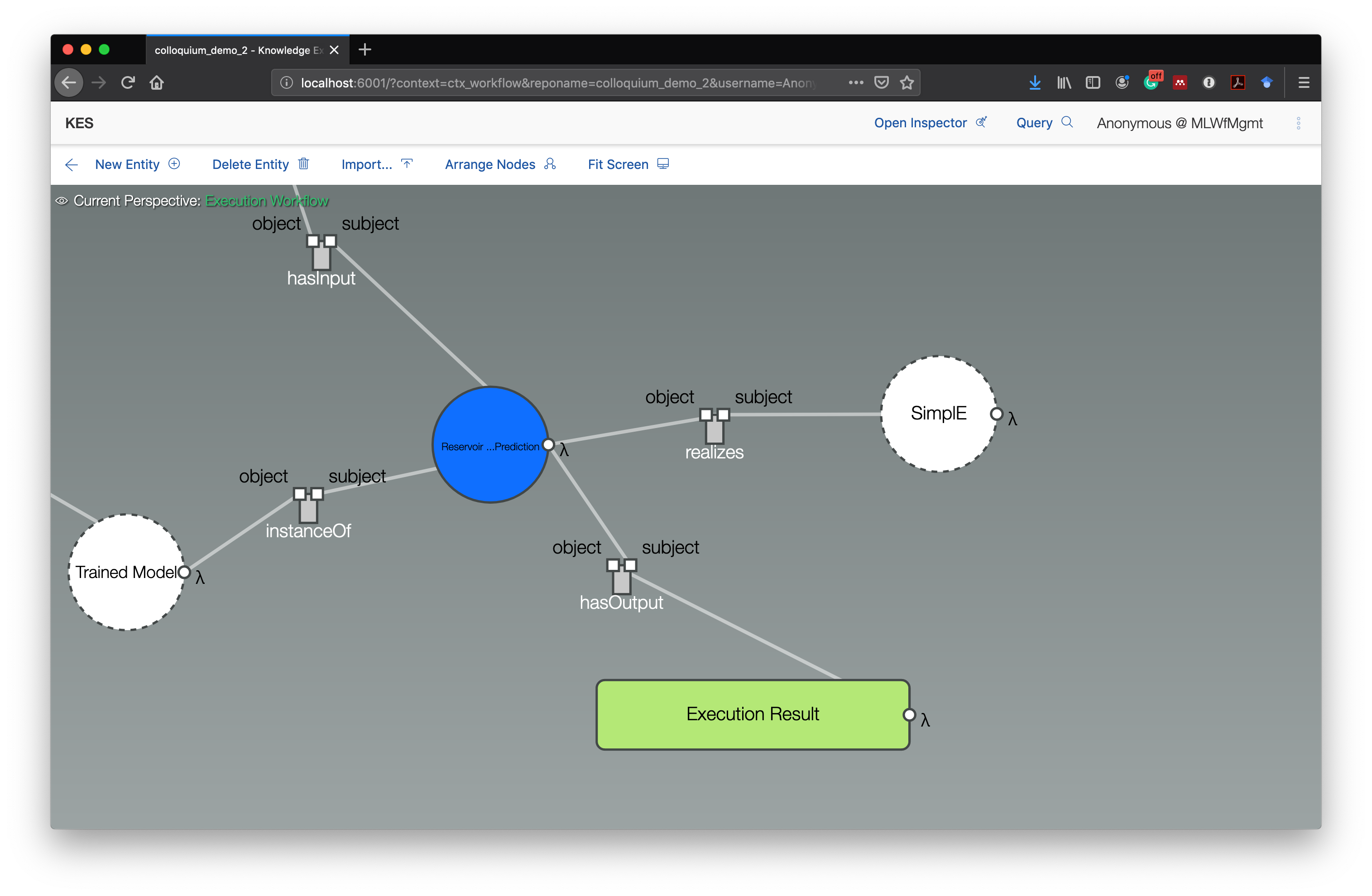}}%
    \caption{Example of execution of a machine learning workflow's component represented in Hyperknowledge framework on KES tool.}
    \label{fig:component-execution}
\end{figure}

The Figure~\ref{fig:component-execution} depicts the KES~(Knowledge Explorer System)~\cite{moreno2018,moreno2018a}, the Hyperknowledge visualization tool. Through KES, the user can visualize the symbolic representation of a MLWf stored in a Hyperknowledge base as well as curate it by adding, updating, or deleting the MLWfs and their components. Also, the KES leverages the third advantage of the MLWfM technique: the execution of MLWfs' components. Through KES, after selecting a component retrieved by a query (Figure~\ref{fig:component-execution}.a), the user is able to execute the component, while the result of this execution is further stored and represented within the MLWf's context (Figures~\ref{fig:component-execution}.b and~\ref{fig:component-execution}.c).

\section{Use Case and Validation}

\subsection{Use case}\label{sec:use-case}

As mentioned before, the proposed Machine Learning Workflow Management gives users the ability to query, create, execute, and share standardized workflows in the problem level of abstraction. To illustrate these properties, we implement a common task in the O\&G Exploration workflow, which is the interpretation of seismic data. We apply machine learning workflows based on image semantic segmentation to identifying the surface separating two different rock layers~\cite{herron2011first}. The seismic information about a specific region provides a view of the general organization of the surfaces in underground strata. The post-stack segmentation consists of identifying the surface separating two different rock layers, also called reflectors~\cite{herron2011first}. In combination with other techniques, such a process aims to identify geological layers and other structures that may lead to potential hydrocarbon deposits~\cite{146848}.

We further extended the ML Schema model with a simple domain ontology to support our use case in the post-stack semantic segmentation problem (see~Figure~\ref{fig:full-ontology} green concepts). The connection between both ontologies is given by characterizing the Seismic concept as a kind of Data and semantic segmentation concept as a kind of task. Finally, we defined more specific ML concepts that implement the post-stack semantic segmentation (see~Figure~\ref{fig:full-ontology} yellow concepts). Based on that, we can carry out MLWfM in the domain, as described in the Section~\ref{sec:validation}.

\subsection{Validation\label{sec:validation}}

To validate the process of MLWfM, we explore its abilities to answer a series of investigations over the original use case and further modifications. The investigations were developed regarding the use case and domain knowledge represented in Figure~\ref{fig:full-ontology}. Also, for the following examples, assume we have a database of seismic images coming from different regions, on which machine learning tasks have been already applied so that previous components can be reused in new workflows, which answers investigations. The investigations were modeled as queries in HyQL, in which we also present their description and the obtained results by evaluating them.

\investigation[Investigation 1]{Which are the trained machine learning models able to perform the semantic segmentation task on seismic images similar to the SeismicA using a similarity factor of at least $0.9$?}

This investigation intends to identify a machine learning model that can pick horizons from an unseen seismic image,~\ie{}, new seismic image (called \emph{SeismicA} in the query). In terms of MLWfM, we are interested in showing the identification and retrieval of the model component of a MLWf based on a specific task that uses a similar input to the queried seismic. The HyQL query used to answer this investigation can be formulated as follows:
\newpage
\begin{mdframed}
\hyql{LET x = \{\\
\tab GET\\
\tab[1cm]Seismic\\
\tab WHERE\\
\tab[1cm]Run hasOutput Model AND Run achieves SeismicSemanticSegmentation AND\\
\tab[1cm]Run hasInput Seismic AND similarSiesmic(SeismicA, Seismic) > 0.9\\
\}\\
SELECT\\
\tab Model\\
WHERE\\
\tab Run hasOutput Model AND Run hasInput x
}
\end{mdframed}

This query retrieves any available machine learning model from a MLWf developed to perform the seismic semantic segmentation task and, also, that consumes seismic images similar to the unseen seismic. Here, the query uses a predefined similarity function that evaluates the similarity between two seismic images, filtering its results by considering a similarity factor greater than $0.9$.

\investigation[Investigation 2]{Which are the trained machine learning models able to perform the seismic semantic segmentation task on seismic images from the same basin as the basin of the seismic image SeismicA?}

This investigation is analogous to the first one, relying on the identification of a machine learning model able to perform the seismic semantic segmentation from an unseen seismic image (again, called \emph{SeismicA} in the query), relying on the assumption that models trained on seismic images from the same basin of SeismicA should also be useful to process SeismicA. In terms of MLWfM, we show the use of domain knowledge to restrict the query's output. The query uses the domain knowledge to specify the relationship between the queried seismic image (\emph{SeismicA}) and other seismic images used in the context of the model's MLWf. The HyQL query used to answer this investigation can be formulated as follows:

\begin{mdframed}
\hyql{LET x = \{\\
\tab GET\\
\tab[1cm]Seismic\\
\tab WHERE\\
\tab[1cm]Run hasOutput Model AND Run achieves SeismicSemanticSegmentation AND\\
\tab[1cm]Run hasInput Seismic\\
\}\\
SELECT\\
\tab Model\\
WHERE\\
\tab Run hasOutput Model AND Run hasInput x AND\\
\tab x hasBasin Basin AND SeismicA hasBasin Basin
}
\end{mdframed}

This query retrieves any available machine learning model from an MLWf developed to perform the seismic semantic segmentation task, and that consumes seismic images of the same basin as the unseen seismic. The presented query uses domain knowledge to relate two seismic images.

\investigation[Investigation 3]{Which are the trained machine learning models able to retrieve horizons with the accuracy of at least $0.85$ from seismic images of Santos Basin?}

This investigation aims to retrieve a trained machine learning model to identify horizons on seismic images from Santos Basin. On the MLWfM aspect, the investigation uses metadata from components to achieve the investigation's goal. Thus, the following query retrieves the set of machine learning models, which general accuracy is above $0.85$, able to identify horizons from the seismic images of the Santos Basin. The HyQL query formulated to answer this investigation:

\begin{mdframed}
\hyql{SELECT\\
\tab Model\\
WHERE\\
\tab Run hasInput Dataset AND Dataset hasBasin SantosBasin AND\\
\tab Run hasOutput Model AND Model hasQuality ModelCharacteristic AND\\
\tab ModelCharacteristic.output=Horizon AND\\
\tab Run hasOutput ModelEvaluation AND ModelEvaluation.accuracy > 0.85
}
\end{mdframed}

This third query retrieves any available machine learning model from an MLWf developed to retrieve horizons from seismic images of the Santos Basin. Here, the query filters the retrieved results by considering models with accuracy greater than $0.85$.

\section{Experimental Evaluation}

In order to evaluate the feasibility of the MLWfM process, we first introduce a new dataset for machine learning workflows and, then, evaluate response time of the proposed technique in loading and querying in the dataset.

\subsection{MLWfD-31k: a new dataset of machine learning workflows\label{sec:dataset}}

The \emph{MLWfD-31k} is a dataset composed of machine learning workflow executions and related information, such as execution dataset, and implementation, specified as instances and relationships of ML Schema concepts and relations. As far as we know, MLWfD-31k is the first dataset for machine learning workflow. It includes different aspects of machine learning models and their execution workflows that allow the performance evaluation of the loading and retrieval capabilities under the Hyperknowledge infrastructure. Additionally, we state that the dataset can also be used as a resource for training and evaluating machine learning models and other frameworks that can recommend the use of MLWf components in different situations that they were initially designed for. The dataset is currently open-source and available at https://github.com/ibm-hyperknowledge/mlwfm/.

The dataset was generated based on the Hyperknowledge model of the ML Schema vocabulary described in Fig.~\ref{fig:full-ontology}. The source data was obtained from PaperWithCode\footnote{paperswithcode.com/} website. The process of acquiring raw data from the website initially resulted in a total of $31256$ entities, summing up to $89319$ triples, describing machine learning contests, their models, and benchmarks. After processing the raw data, \ie{}, converting entities into Hyperknowledge nodes and their relationships into Hyperknowledge links and connectors, according to the Hyperknowledge model of the ML Schema vocabulary, the entities were divided into $15$ concepts, as seen in Table~\ref{tab:ds-content}. The process resulted in a total of $17733$ Hyperknowledge nodes, $71736$ Hyperknowledge links, and $12$ Hyperknowledge connectors.

\begin{table}[t]
\centering
\begin{tabular}{@{}cc@{}}
\toprule
\textbf{Concept}                      & \textbf{Number of instances} \\\midrule
Area                         & 17       \\
Subarea                      & 553      \\
Task                         & 1097     \\
Dataset                      & 615      \\
DatasetCharacteristic        & 615      \\
Data                         & 615      \\
DataCharacteristic           & 615      \\
Model                        & 3185     \\
Run                          & 3187     \\
ModelCharacteristic          & 3185     \\
Algorithm                    & 3185     \\
Implementation               & 3186     \\
ImplementationCharacteristic & 3182     \\
ModelEvaluation              & 3186     \\
EvaluationMeasure            & 5448     \\\midrule
\textbf{Total}                        & 31256 \\\toprule
\end{tabular}
\caption{MLWfD-31k's entities distribution per concept.}
\label{tab:ds-content}
\end{table}

\subsection{Response time evaluation}

In order to demonstrate the feasibility of the proposed framework in Hyperkowledge, we evaluated it in some common MLWf activities present in the MLWfD-31k dataset. We analyzed dataset load time in the Hyperknowledge Base and response times of five queries constructed based on the ML Schema vocabulary and coded in HyQL on the test dataset.

For this evaluation, we employed a single machine running a Fedora 31 (Workstation Edition) operational system. The machine comprises an Intel Core i7-8700T CPU, an 8 GB DDR4 RAM, and 500 GB Toshiba MQ01ACF0 hard disk. The Hyperknowledge Base infrastructure was deployed in this machine using the Docker framework version 19.03.5. For these experiments, the Hyperknowledge Base used the MongoDB version 4.2.5 as storage.

\begin{figure}[t]
  \subfloat[]{
	\begin{minipage}[c][1\width]{
	   0.35\textwidth}
	   \centering
	   \includegraphics[width=\textwidth]{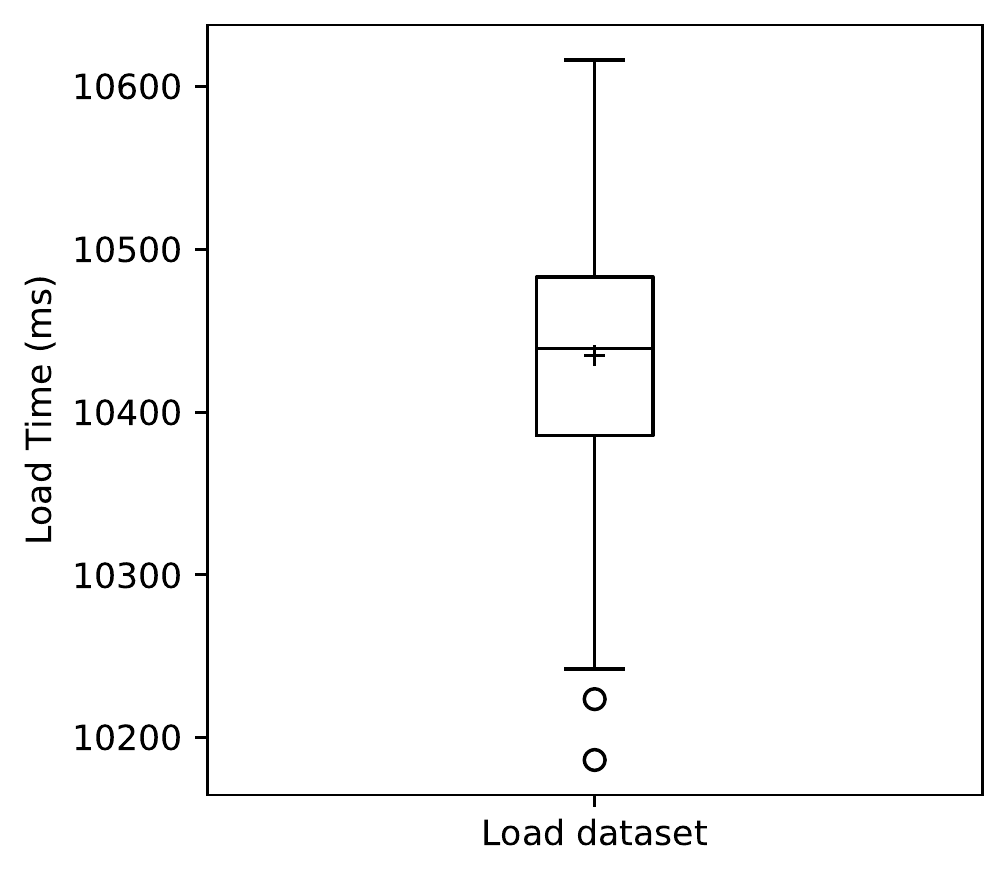}
	\end{minipage}}
 \hfill 	
  \subfloat[]{
	\begin{minipage}[c][1\width]{
	   0.35\textwidth}
	   \centering
	   \includegraphics[width=\textwidth]{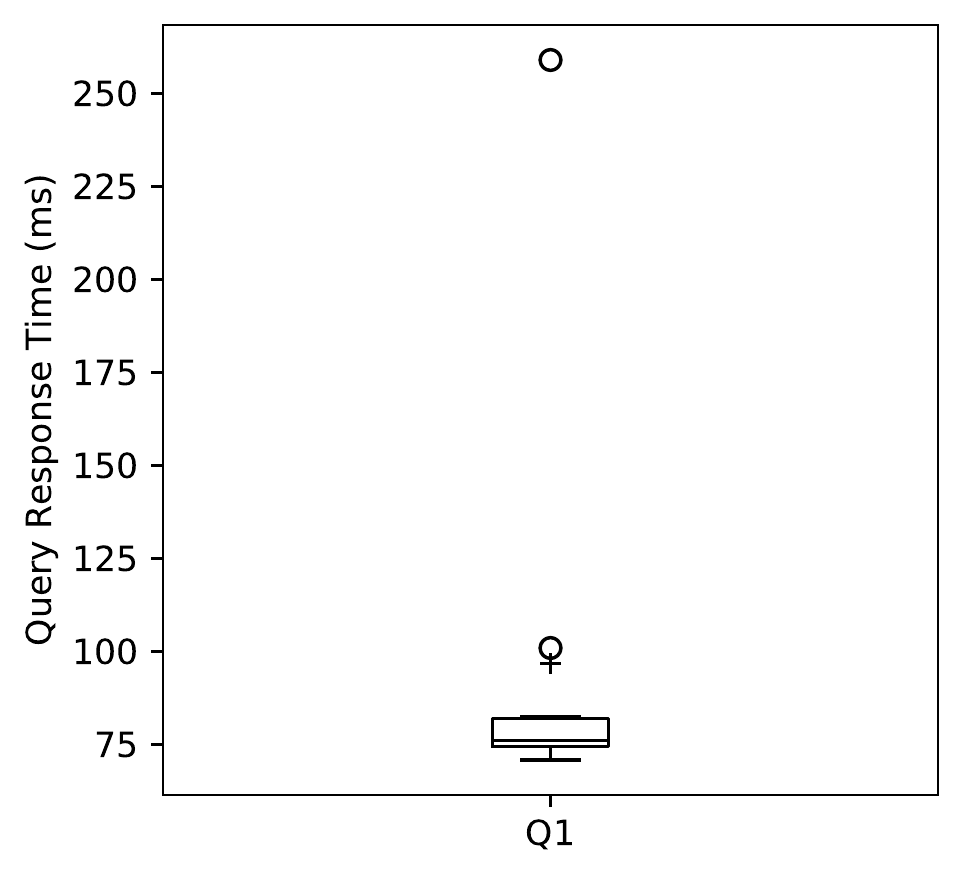}
	\end{minipage}}
 \hfill	
  \subfloat[]{
	\begin{minipage}[c][1\width]{
	   0.35\textwidth}
	   \centering
	   \includegraphics[width=\textwidth]{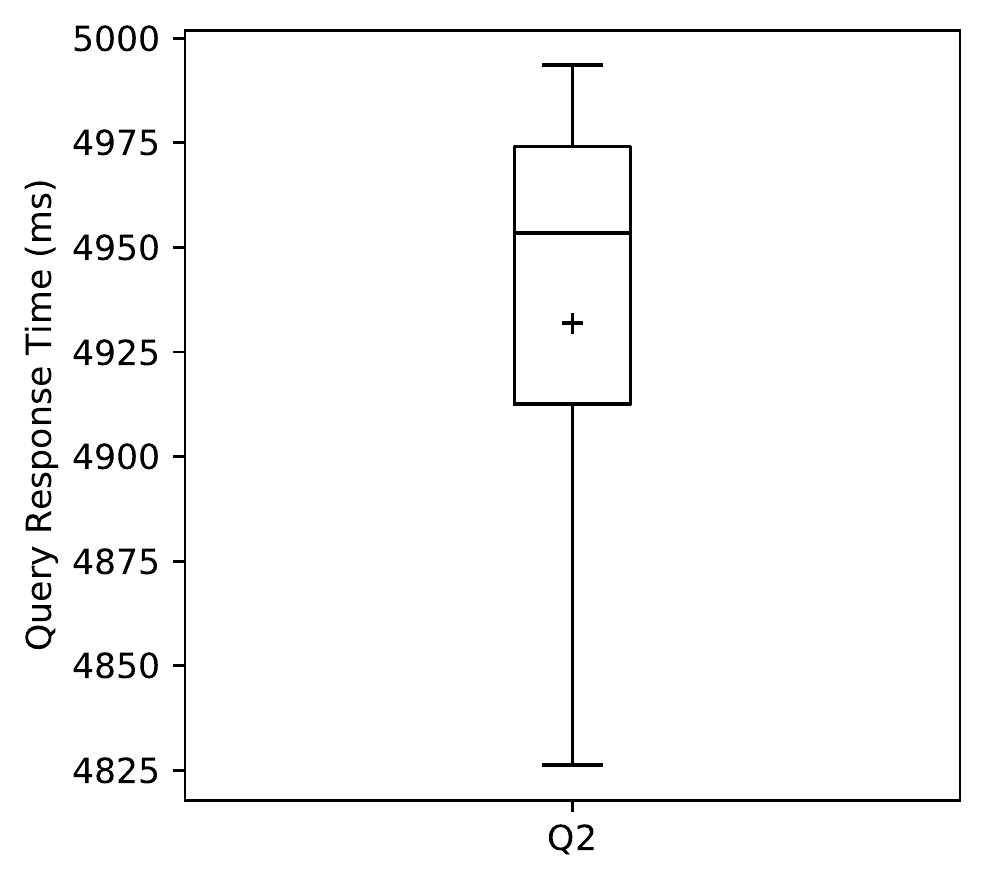}
	\end{minipage}}
	\hfill	
	\subfloat[]{
	\begin{minipage}[c][1\width]{
	   0.35\textwidth}
	   \centering
	   \includegraphics[width=\textwidth]{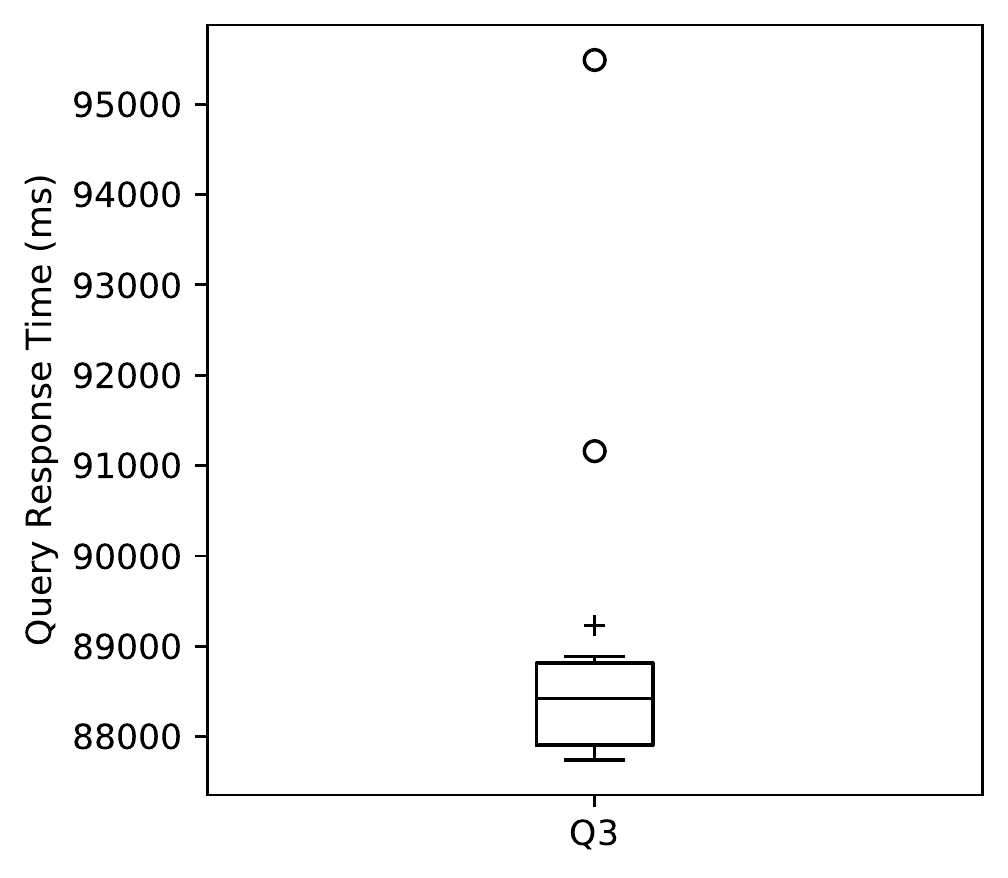}
	\end{minipage}}
 \hfill 	
  \subfloat[]{
	\begin{minipage}[c][1\width]{
	   0.35\textwidth}
	   \centering
	   \includegraphics[width=\textwidth]{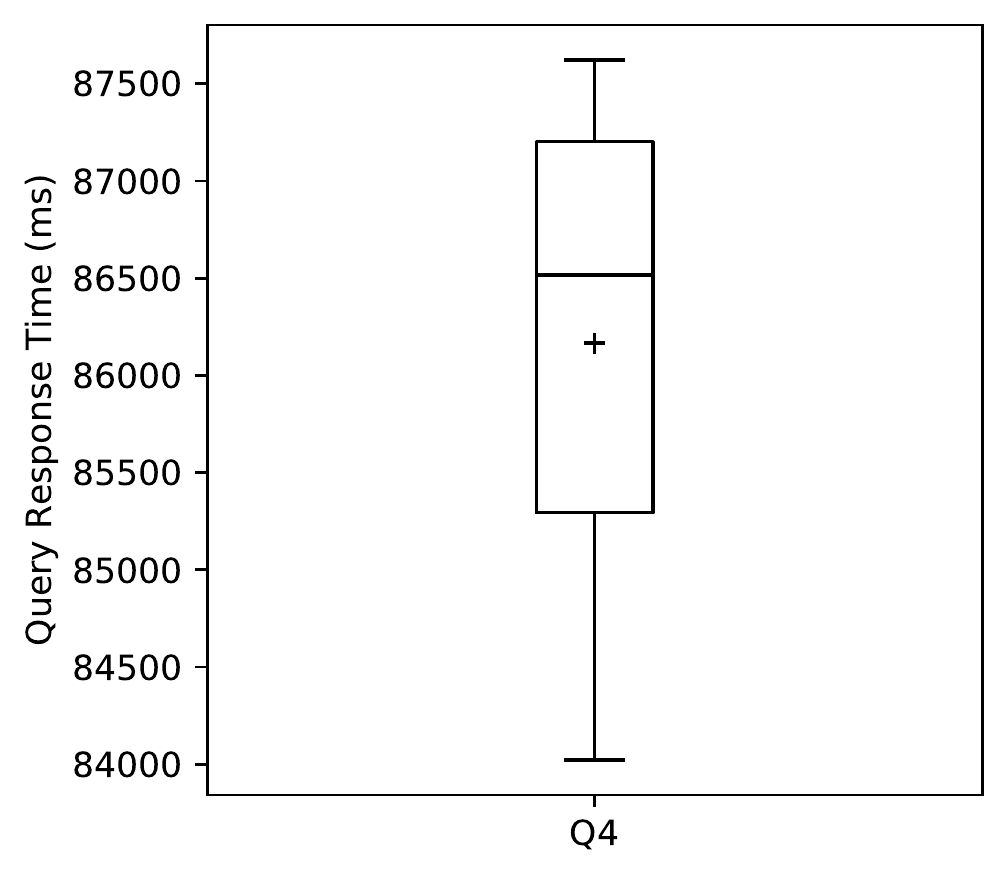}
	\end{minipage}}
 \hfill	
  \subfloat[]{
	\begin{minipage}[c][1\width]{
	   0.35\textwidth}
	   \centering
	   \includegraphics[width=\textwidth]{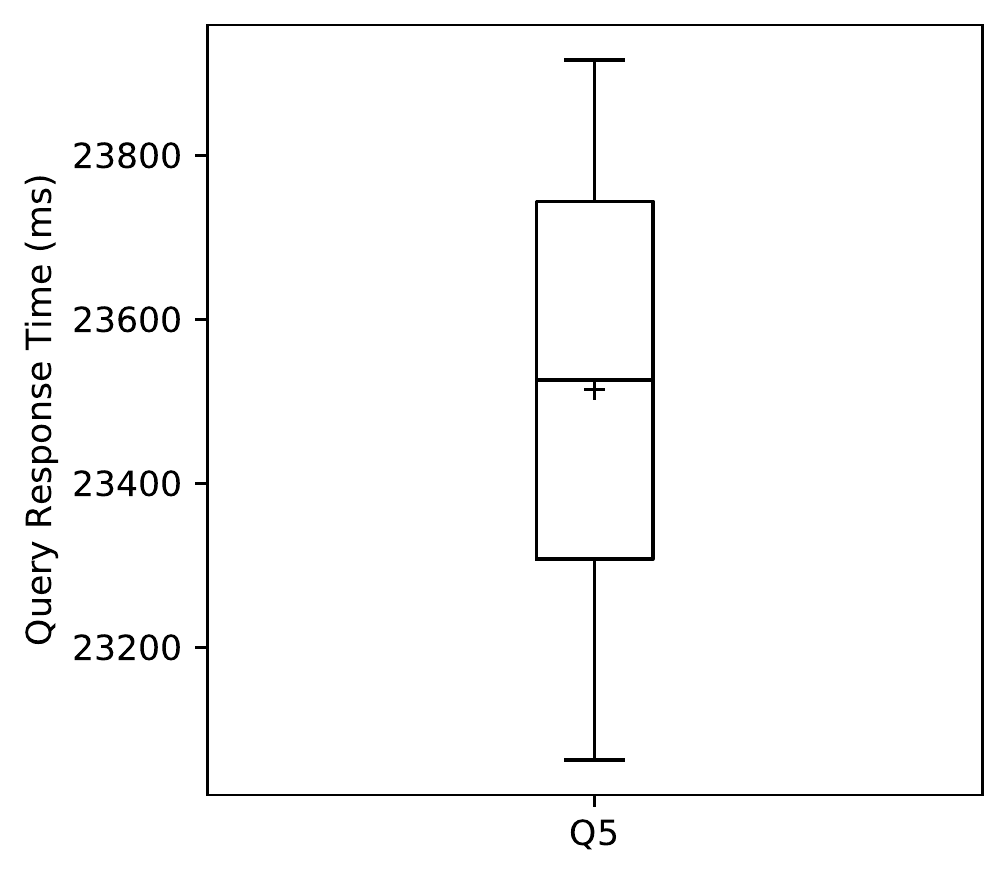}
	\end{minipage}}\\
\caption{Response time evaluation. (a) illustrates the dataset's loading time, and (b)-(e) the response time of each evaluation query, all regarding the implementation of MLWfM over Hyperknowledge and HyQL. The circles ($\circ$) and the crosses ($+$) show the distributions' outliers, and the distributions' means, respectively.}
\label{fig:evaluation}
\end{figure}

Table~\ref{tab:queries} shows the five queries used in the evaluation. These queries were chosen based on their relevance for retrieval and reuse of ML components within the dataset domain, as well as its complexity by analyzing the number of what we call \emph{link patterns}, \ie{}, patterns that relates two entities or concepts in the HyQL query. For instance, an example of two link patterns in a query is \hyqlinline{Run achieves Task AND Run hasOutput Model}, where \emph{Run achieves Task} is the first link pattern and \emph{Run hasOutput Model} is the second one. The analyses further discussed in this Section are based on the data depicted in Figure~\ref{fig:evaluation}. These data were gathered after performing each query as well as the dataset loading for 100 times.

\begin{table}[]
\begin{tabularx}{\textwidth}{lX}
\toprule
\textbf{$Q_1$}: & \hyqlinline{SELECT Subarea WHERE Subarea hasTask object\_detection} \\\midrule
\textbf{$Q_2$}: & \hyqlinline{SELECT Run WHERE Run hasInput Data and Data.id=pascal\_voc\_2012} \\\midrule
\textbf{$Q_3$}: & \hyqlinline{SELECT Implementation WHERE Implementation implements Algorithm AND Run realizes Algorithm AND Run achieves semantic\_segmentation} \\\midrule
\textbf{$Q_4$}: & \hyqlinline{SELECT Model WHERE Run hasOutput Model AND Run hasOutput ModelEvaluation AND ModelEvaluation specifiedBy accuracy} \\\midrule
\textbf{$Q_5$}: & \hyqlinline{SELECT Algorithm WHERE Run realizes Algorithm AND Run achieves object\_detection AND Run achieves unsupervised\_image\_classification AND Run hasInput Data AND Data.id=imagenet\_detection} \\
\toprule
\end{tabularx}
\caption{Evaluation queries}
\label{tab:queries}
\end{table}

Specifically, Fig.~\ref{fig:evaluation}.(a) references the dataset load time results obtained from the executions. On average, the MLWfM implemented over the Hyperknowledge infrastructure spends $10$ seconds to load the dataset in the Hyperknowledge Base.
Moreover, figures Fig.~\ref{fig:evaluation}.(b)-(f) refer the queries response time.
The first query (Fig.~\ref{fig:evaluation}.(b) and Table~\ref{tab:queries}.$Q_1$) retrieves from the dataset all entities related to the \emph{Subarea} concept, that have \emph{object detection} as an associated task. This query uses one link pattern and has a median response time of $76$ milliseconds.
The second query (Fig.~\ref{fig:evaluation}.(c) and Table~\ref{tab:queries}.$Q_2$) retrieves all executions, \ie{}, entities instances of \emph{Run} concept, that have as an input the \emph{Pascal VOC 2012}\footnote{http://host.robots.ox.ac.uk/pascal/VOC/voc2012/} dataset. This query has one link pattern and one access to the entity's property \emph{id}. It is considerably slower when compared with the first one, once it evaluates $3187$ different executions over different datasets $615$.
The third (Fig.~\ref{fig:evaluation}.(c) and Table~\ref{tab:queries}.$Q_3$) and fourth (Fig.~\ref{fig:evaluation}.(d) and Table~\ref{tab:queries}.$Q_4$) evaluated queries use three link patterns and are the slowest queries presented spending, as they are computed over the concepts with the highest numbers of instances. The third query searches on the dataset for implementations associated with executions that accomplish the semantic segmentation task, spending up to $87$ seconds on its best case. The fourth query retrieves machine learning models that have their training performance evaluated using the accuracy metric. It has spent, on average, $86$ seconds to be evaluated.
Finally, the fifth query evaluated (Fig.~\ref{fig:evaluation}.(e) and Table~\ref{tab:queries}.$Q_5$) have the greatest number of link patterns, four in total, and one access to the \emph{Data}'s entities property \emph{id}. Semantically, the query retrieves algorithms that are used in executions that perform \emph{object detection} or \emph{unsupervised image classification} tasks over the \emph{imagenet}\footnote{http://www.image-net.org/} dataset. On average, this query spends $23$ seconds to be performed, which is reasonable response time.

These queries are examples of real use of components' retrieval. They allow the users (\eg{}, scientists, machine learning engineers) to reuse the components in different scenarios, such as new MLWfs creation. For instance, the fifth query unifies algorithms initially developed for two different tasks, but they are able to be performed interchangeably. Besides, all queries presented acceptable response times to be used as an application.

\section{Conclusion and Future Work}

In this work, we define and address the machine learning workflow management as a technique for symbolic representation, execution, and creation of machine learning workflows. We introduce the symbolic representation in the context of Hyperknowledge framework. We propose the retrieval of MLWfs' components based on the Hyperknowledge Query Language and the composition of these retrieved components to create a new MLWf. We show KES as a tool to support the execution of components. We validate our approach demonstrating scenarios in the Oil \& Gas industry, exemplifying our technique's capabilities. Finally, we propose MLWfD-31k, a new dataset to evaluate our technique, but can also be used in similar tasks in similar problems. Also, we show the experimental evaluation of MLWfM using the Hyperknowledge infrastructure and the proposed dataset. Potential future works include the composition of new machine learning models using fragments from existing models~\cite{morenopatent2019}.

\bibliographystyle{unsrt}  
\bibliography{refs}

\end{document}